\documentclass{nature}

\usepackage[utf8]{inputenc}
\usepackage{graphicx}
\usepackage{amsmath}
\usepackage{amssymb}
\usepackage{endnotes}
\usepackage{url}
\usepackage[percent]{overpic}

\usepackage{booktabs}
\usepackage{longtable}
\usepackage{array}
\usepackage{verbatim}

\usepackage[resetlabels]{multibib}
\newcites{Methods}{Methods-Only References}

\bibliographystyleMethods{naturemag}
\continuouslabelstrue

\title{A Data-free Universal Prior over Syntactic Structures}

\author{Ferm\'{\i}n Moscoso del Prado Mart\'{\i}n$^{1}$}

\begin{document}

\maketitle

\begin{affiliations}
 \item Department of Computer Science and Technology, University of Cambridge, United Kingdom
\end{affiliations}

\begin{abstract}
The probabilities of syntactic structures in human languages are assumed to  emerge fully from language-specific experience. Here, I show that a universal prior over syntactic structures emerges from a model of human language production, in which words are progressively integrated into syntactic structure. Without fitting any parameters to specific language data, the resulting prior assigns higher probabilities to attested than to random dependency trees in all 138 typologically diverse languages examined. These prior probabilities correlate positively with those estimated from corpora in 33 of 34 languages. The results indicate that part of the probability structure of syntax can arise independently of language-specific learning. This identifies human language production as a possible cognitive source of universal statistical structure in language, while providing a data-independent structural bias for probabilistic models, including large language models.
\end{abstract}

Language is fundamentally probabilistic. Humans continuously estimate the probabilities of forthcoming words and syntactic structures during language comprehension and production, and acquire these expectations from finite linguistic experience.\cite{Shannon:1951,Hale:2001,Levy:2008,Jaeger:2010} Likewise, modern large language models represent language as probability distributions learned from vast corpora, enabling them to predict and generate remarkably fluent text.\cite{Vaswani:2017,Brown:2020} Understanding the origins of these probabilities is therefore central to both cognitive science and artificial intelligence.

Probabilistic theories of syntax estimate the probabilities of syntactic structures from language-specific observations.\cite{Johnson:1998,Klein:2003} A fundamental question is whether, across languages, there exists a universal prior over syntactic structures, or whether all structures are equally plausible before linguistic evidence is observed. Universal hierarchical priors have proved invaluable in Bayesian nonparametrics.\cite{Neal:2003,Teh:2006,Knowles:2011} A cognitively grounded universal prior would provide a language-independent starting point from which language-specific probabilities could be learned, separating universal structural preferences from those acquired through experience. It could also provide a principled inductive bias for probabilistic models of language, including large language models.

In this paper, I show that a model of incremental human language production\cite{Moscoso:2026} defines a universal data-free prior distribution over syntactic structures. This prior predicts both which structures occur across languages and their probabilities estimated from language-specific data.

\begin{figure}[t]
  \includegraphics[width=\columnwidth]{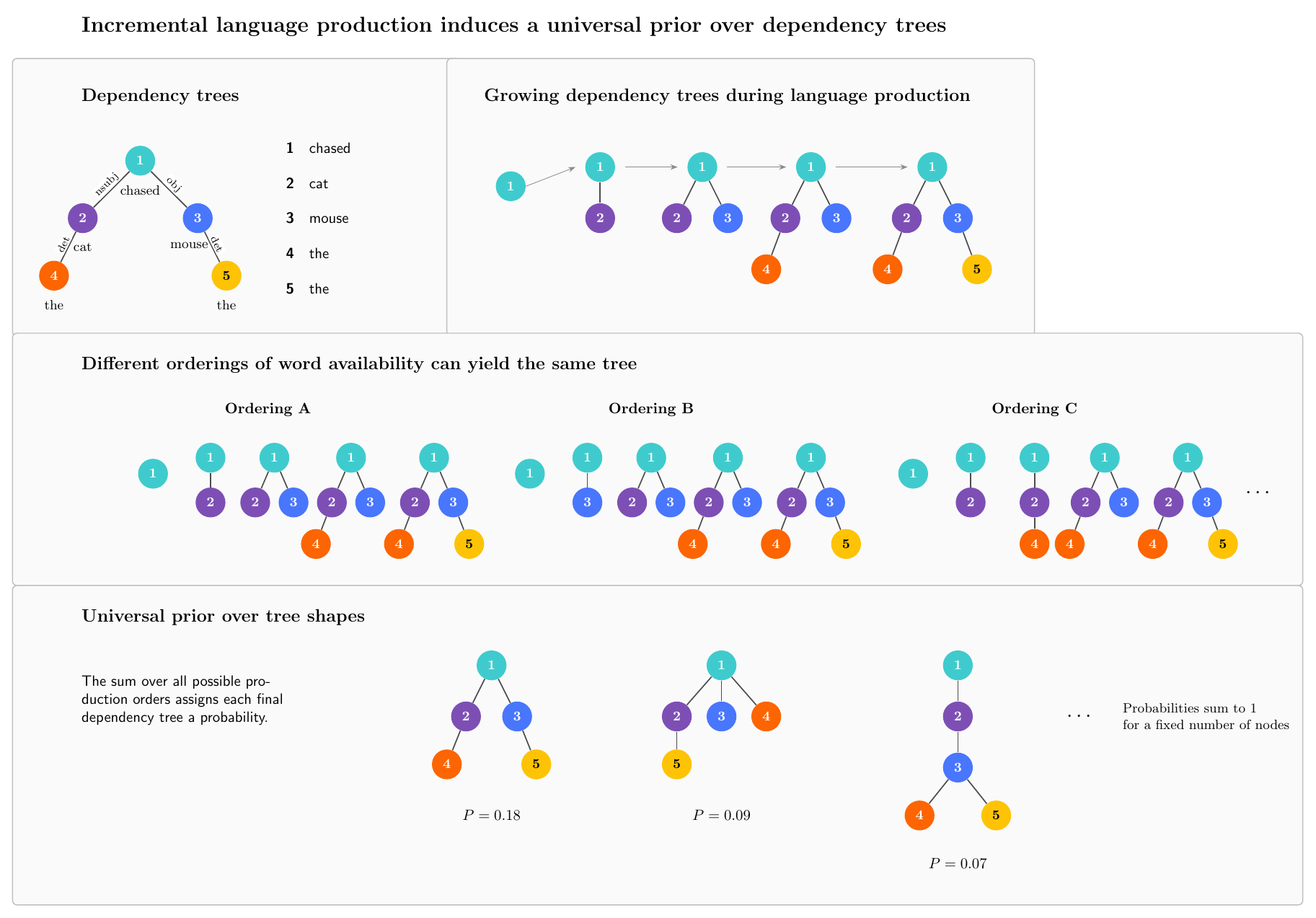}
\caption{\textbf{Incremental language production induces a universal prior over dependency trees.}
A dependency tree can be constructed incrementally as words become available during language production. Different orders of word availability can generate the same final dependency tree. Summing across all possible production orders assigns a probability to each dependency-tree shape, thereby defining a prior distribution over trees of a fixed number of nodes. The numerical probabilities shown are just for illustration.}
\label{fig:schema}
\end{figure}

\section*{A universal prior over syntactic structures}

Language production is inherently incremental, with syntactic
structure emerging progressively during the planning and production
of an utterance rather than being assembled all at once.\cite{Levelt:1989,Ferreira:2007}
More generally, constraints imposed by language production have been
argued to shape the structure of language itself.\cite{MacDonald:1999,MacDonald:2013}
At every stage of production, only a partial dependency tree has been
constructed, and each newly available word must be integrated into
this evolving structure by establishing a syntactic dependency with
one of the words already present.
This process can be naturally formulated as a growing network, in
which words correspond to nodes, syntactic dependencies to edges,
and the partial dependency tree grows through the successive addition
of new nodes.\cite{Moscoso:2026}
Fig.~\ref{fig:schema} illustrates this generative process.

The choice of where to attach each newly available word depends on
multiple syntactic, semantic, pragmatic, and contextual factors that
are difficult to model explicitly.
The current partial dependency tree nevertheless provides observable
information about this latent attachment propensity: The number of direct dependents already attracted by a word provides a natural observable proxy for its latent propensity to attract further ones.\cite{Moscoso:2026}
Therefore, a preferential attachment, in which
attachment probability increases with the number of existing
connections, can provide an adequate model of this process.\cite{Barabasi:1999} However, language production is only approximately incremental.
Speakers often plan multiple words or constituents simultaneously,
and previously activated syntactic structures facilitate the
production of subsequent ones through structural priming.\cite{Bock:1986,Pickering:2008,Mahowald:2016}
These interactions weaken the assumption that each attachment
decision is made independently, attenuating the cumulative
advantage characteristic of linear preferential attachment.
The attachment process is therefore expected to follow
sublinear preferential attachment (sPA),\cite{Krapivsky:2000}
in which the probability that a newly available word attaches
to node $i$ at time $t$ is
\begin{equation}
P_t(i)\propto (k_t(i)+1)^\alpha,
\qquad 0<\alpha<1,
\label{eq:spa}
\end{equation}
where $k_t(i)$ denotes the number of direct dependents (i.e., the out-degree) of node $i$
in the partially constructed dependency tree, and $\alpha$ is a parameter indicating the degree to which the process is purely incremental.

Different node insertion orders may lead to the same final structure.
Among all node orderings, only those in which every node appears after all of its
ancestors are compatible with incremental construction of a specific observed tree.
Denoting the set of compatible orderings for a tree $T$ by $V(T)$,
the probability of a dependency tree is obtained by summing
over all compatible orderings,
\begin{equation}
P(T)=
\sum_{o\in V(T)}
P(o)\,
P(T\mid o).
\label{eq:treeprob}
\end{equation}
Assuming no prior knowledge beyond the sentence length,
all $n!$ node orderings should --a priori-- be considered equally likely.
The tree probability therefore decomposes into
\begin{equation}
P(T)=
\frac{|V(T)|}{n!}
\left<P(T\mid o)\right>_{o\in V(T)},
\label{eq:decomposition}
\end{equation}
where $|V(T)|$ is the number of compatible node orderings and
$\left<P(T\mid o)\right>_{o\in V(T)}$
is the average probability with which the tree is generated
across those node orderings.
The compatible node orderings are precisely the
\emph{increasing labellings} (equivalently,
the linear extensions) of the tree,
a classical combinatorial object with known cardinality.\cite{Knuth:1998}

The {sPA} process induces a
prior probability over dependency trees. Rather than fixing the attachment parameter $\alpha$ of Eq.~\ref{eq:spa}, I treat it as a nuisance parameter that can be marginalised away by integrating over its full sublinear range with an uninformative prior (see Methods).\endnote{The results are not dependent on the specific prior or on marginalisation over $\alpha$: Repeating the analyses produces qualitatively equivalent results either using alternative priors or just fixing it to the previously estimated optimal value of $\alpha=.42$.\protect\cite{Moscoso:2026}} The resulting tree probability is therefore
\begin{equation}
P(T)
=
\frac{|V(T)|}{n!}
\int_0^1
\left\langle P(T\mid o,\alpha)\right\rangle_{o\in V(T)}
p(\alpha)\,d\alpha .
\label{eq:integration}
\end{equation}
This results in a data-independent prior distribution over
dependency trees. The computation of the number of compatible orderings and
the numerical estimation of this integral is described in the Methods.

\section*{The universal prior distinguishes language from randomness}
I evaluated the universal sPA prior on up to 500 dependency
trees randomly sampled from each of 138 languages in the Universal
Dependencies v2.11 corpora,\cite{deMarneffe:etal:2021} retaining all
suitable trees for languages with fewer than 500.
I considered sentences containing 50 words or fewer
(see Methods for corpus preprocessing and sampling, and Extended Data Table~1 for the list of languages).

For each attested tree of $n$ words, I compared its probability
under the sPA prior with that under a uniform prior over rooted
labelled trees.
By Cayley's formula,\cite{Cayley:1889} there are $n^{n-2}$ labelled trees and
$n$ possible choices of root, yielding $n^{n-1}$ rooted labelled
trees and hence a uniform probability of $n^{1-n}$ for each tree.
As sanity checks, I used two controls to verify that this comparison
discriminates between the two generative processes:
For each attested tree, I generated a tree of the same size from
a true sPA model and another uniformly from the
space of rooted labelled trees, and evaluated both under the same
two priors.

\begin{figure}[t]
\centering
\begin{tabular}{lll}
{\bf a} & \multicolumn{2} {l} {\bf b} \\
  \includegraphics[width=0.3\linewidth]{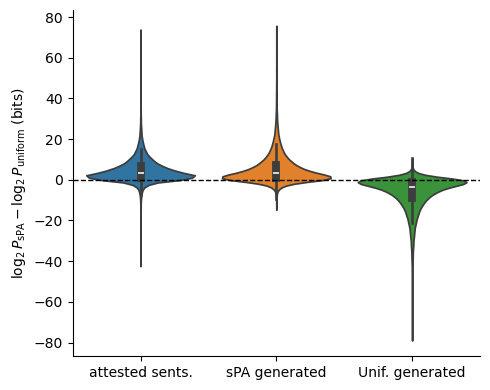} &
  \multicolumn{2} {l} {\includegraphics[width=.6\linewidth]{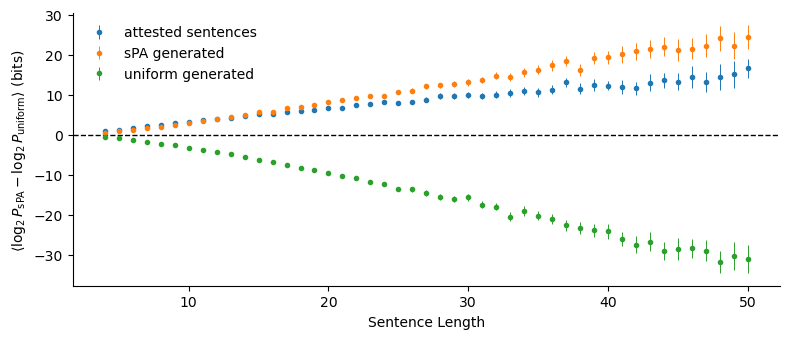}} \\
\multicolumn{3} {l} {\bf c} \\
\multicolumn{3} {l} {\includegraphics[width=.9\linewidth]{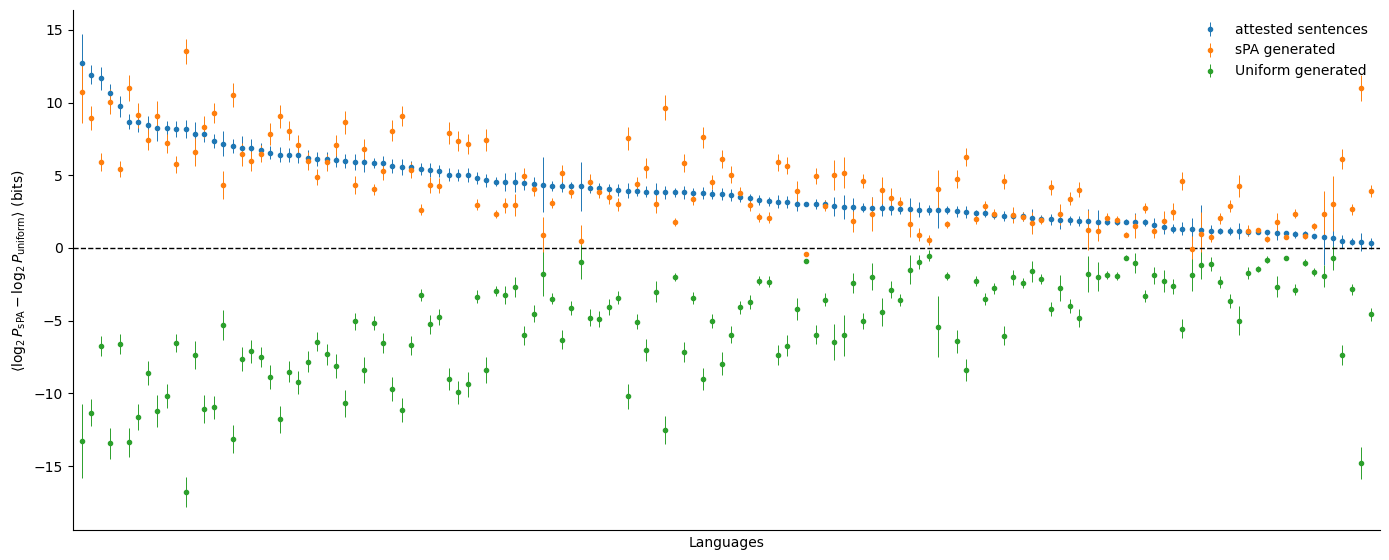}}
 \end{tabular}
\caption{{\bf 
Probability advantage of the sPA model over the uniform model.}
{\bf a}, Distribution of $\log_2 P_{\rm sPA}-\log_2 P_{\rm uniform}$
for attested dependency trees, trees generated by the sPA model, and
uniformly generated random trees. Violins show the distributions;
internal boxes indicate the median and interquartile range. {\bf b}, Mean log-probability difference for each attested sentence length, shown for
the same three sets of trees. Points indicate means and error bars
95\% confidence intervals.  {\bf c}, Mean log-probability difference for each language, shown for the same three sets of trees. Points indicate means and error bars 95\% confidence intervals. Languages are ordered by the mean difference for attested sentences. In all three panels, positive values indicate higher probability under sPA. Conversely, negative values reflect that the probability is higher under the uniform model than under the sPA one.
}
\label{fig:probability}
\end{figure}

The sPA prior assigns systematically higher probabilities to
attested dependency trees than the uniform prior
(Fig.~\ref{fig:probability}a).
As expected, the same preference is observed for trees generated
by the sPA process, whereas uniformly generated trees show the
opposite pattern.
For attested trees, the effect is remarkably consistent across every  sentence length examined (Fig.~\ref{fig:probability}b) and for every language studied (Fig.~\ref{fig:probability}c). Together, these results show that attested dependency-tree topologies are systematically and substantially closer to those favoured by the sPA prior than expected under a uniform distribution over trees, and that this pattern generalises across typologically diverse languages.

\section*{The universal prior predicts language-specific probabilities}

The preceding analysis establishes that attested dependency-tree
shapes are systematically favoured by the sPA prior, but this pattern
could in principle partly reflect conventions used to construct
dependency-tree annotations rather than the probabilistic structure
of language itself.
A stronger test is whether sPA probabilities predict the probabilities
of syntactic structures estimated independently from the statistics
of individual languages.
To approximate these language-specific probabilities, I used
probabilistic context-free grammars (PCFGs).\cite{Johnson:1998} For each language in the Universal Dependencies represented by at least 10,000 non-empty trees ($34$ languages), I induced a PCFG (see Methods for the detailed procedure). From each of the languages, I randomly sampled 500 sentences containing between
4 and 30 words. I computed the universal sPA probability of
each dependency tree and the probability of the corresponding
context-free derivation tree under the language-specific PCFG.

\begin{figure}[t]
\begin{tabular}{lll}
{\bf a} & \multicolumn{2} {l} {\bf b} \\
  \includegraphics[width=0.3\linewidth]{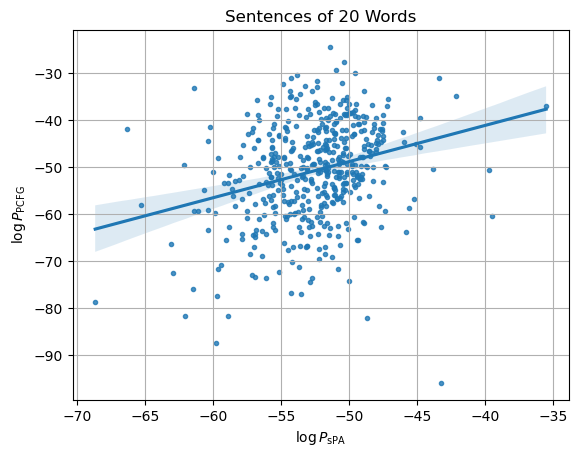} &
  \multicolumn{2} {l} {\includegraphics[width=.6\linewidth]{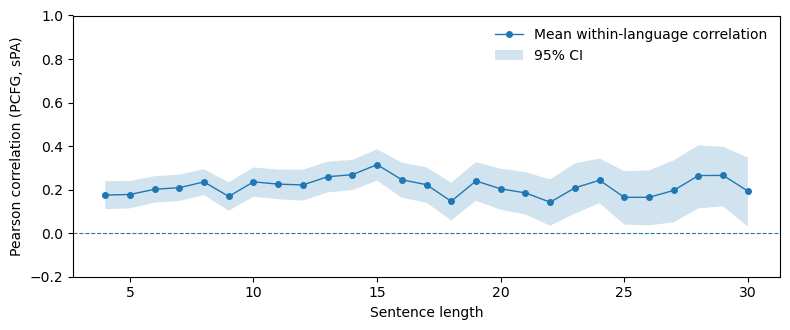}} \\
\multicolumn{3} {l} {\bf c} \\  
  \multicolumn{3} {l} {\includegraphics[width=.9\linewidth]{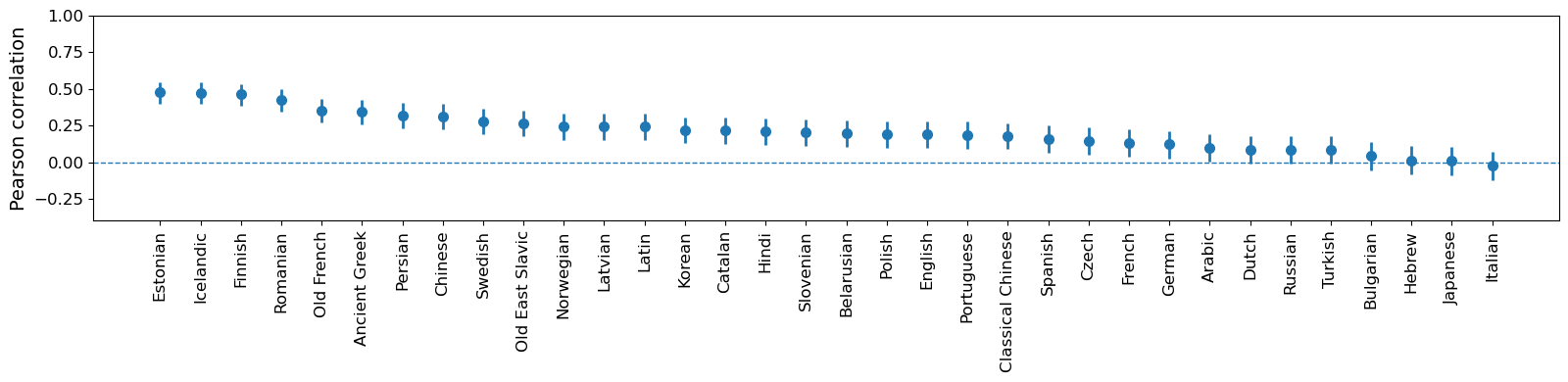}} \\
 \end{tabular}
\caption{{\bf Correlation between sPA and language-specific PCFG
probabilities.}
{\bf a}, Example of the relationship between the two probabilities
for sentences of length 20 across all 34 languages.
Each point represents an individual sentence; the line shows the
linear regression fit, with shading indicating its 95\% confidence interval.
{\bf b}, Mean within-language Pearson correlation as a function of
sentence length, with correlations combined across languages using
Fisher's \(z\) transformation.
Shading indicates 95\% confidence intervals.
{\bf c}, Mean within-language Pearson correlation for each language,
with correlations computed separately by sentence length and combined
across lengths using Fisher's \(z\) transformation.
Error bars indicate 95\% confidence intervals, and languages are
ordered by decreasing correlation.
In {\bf b} and {\bf c}, correlations are weighted by \(n-3\), the
inverse of the approximate sampling variance of the Fisher-transformed
correlation, where \(n\) is the number of sentences contributing to
each correlation.}
\label{fig:correlation}
\end{figure}

The sPA and language-specific PCFG probabilities were positively
correlated at every sentence length examined
(Fig.~\ref{fig:correlation}a,b), with all correlations being
significantly greater than zero, and of a similar magnitude across all lengths, with no clear evidence of a systematic relation between sentence length and correlation strength. The pattern was remarkably consistent across languages
(Fig.~\ref{fig:correlation}c): The correlations were positive for
33 of the 34 languages studied. Italian was the sole exception, with a non-significant slightly negative average correlation. This exception may reflect limitations of the PCFG approximation
rather than a failure of the universal prior: in the preceding
analysis, Italian dependency trees, like those of every other
language, are systematically better predicted by the sPA prior than
by the uniform prior (Fig.~\ref{fig:probability}c). These correlations are only moderate in magnitude. This is, however, substantial 
considering that the sPA prior does not contain any language-specific information. The relationship was further confirmed by a linear mixed-effects
analysis of sentence-level probabilities, accounting for variation
across languages.
The length-standardised sPA probability is a positive
predictor of language-specific length-standardised PCFG probability
($\beta=.171$, 95\% CI $[.142,.199]$, $P<.0001$; see Methods).

\section*{The origins of syntactic structure probabilities}

These results suggest that the probabilities of syntactic structures do not arise solely from statistical regularities acquired through language-specific
experience. The incremental nature of human language production already induces a
non-uniform distribution over possible structures, before any
language-specific statistics are introduced. Linguistic experience can
therefore be understood as updating a structured universal prior rather
than inducing the probabilities of syntactic structures from scratch. The correlation
between the data-free prior and probabilities estimated independently
from individual languages is consistent with this account: A universal
component arising from language production remains detectable within
the distributions shaped by language-specific experience.

Probabilities over syntactic structures are known to influence human language production.\cite{Jaeger:2010} The present results suggest that the reverse also holds: The processes involved in language production themselves help shape those probabilities. Statistical learning may therefore operate on a probability distribution already structured by the process of language production. More generally, the present results show how a universal probabilistic bias can emerge from a cognitive process without requiring either language-specific statistical estimation or an initially uniform distribution over syntactic structures. The same principle could provide a data-independent structural bias for computational models, including large language models, whose probability distributions are predominantly learned from linguistic data.

The idea that the processes generating structures can constrain the forms that emerge echoes a broader principle introduced by D'Arcy Thompson in the study of biological form: Understanding why particular forms arise requires considering the processes by which these forms are grown.\cite{Thompson:1917} Recent cross-linguistic evidence similarly indicates that grammatical variation is constrained by enduring statistical universals, attributed to cognitive and communicative pressures.\cite{Verkerk:2026} The present results suggest that the incremental construction of syntactic structures may constitute one such cognitive pressure. The probabilistic structure of language may therefore reflect not only what speakers have experienced, but also the processes through which language itself is produced.

\section*{References}
\bibliography{prior}

\begin{methods}

\subsection{Corpora and preprocessing}

I analysed all languages represented in the Universal Dependencies
 v2.11 treebanks (UD).\cite{deMarneffe:etal:2021}
For languages represented by multiple treebanks, I concatenated all
treebanks listed for that language. I retained only lexical vertices, removing punctuation and range
vertices (e.g., ``3--4'' in the CoNLL format used in UD), deleted relation labels,
and replaced vertex labels with consecutive integers.
After these modifications, I retained only graphs that remained
actual trees (see Extended Data Table~\ref{tab:languages} for the count of valid trees available in each language). 

For the comparison between the sPA and uniform priors, I randomly sampled up to 500 trees (taking all trees whenever fewer than 500 were available) with a maximum length of 50 words. For the analysis of the correlation between sPA and PCFG probabilities, I instead used a sample of 500 trees with a maximum length of 30 words, as longer trees were increasingly rare.

Because trees with $N\leq3$ do not exhibit any shape variability
(all trees with three or fewer vertices are simultaneously line and
star graphs), in both samples I considered only dependency graphs with at least four vertices.

\subsection{Computing the universal prior}

Computing the universal prior in
Eq.~\ref{eq:integration}
requires two quantities:
the number of node orderings
compatible with a dependency tree,
and the average probability with which
the tree is generated across those orderings.
The former can be computed exactly,
whereas the latter is estimated by
Monte Carlo sampling.

{\bf Compatible node orderings.} In the network growth model, only orderings in which every node appears
after all of its ancestors are compatible
with a given dependency tree.
These compatible orderings are
precisely the increasing labellings
of the tree, whose number is given exactly by the hook length
formula,\cite{Knuth:1998}
\begin{equation}
|V(T)|=
\frac{n!}
{\prod_{v\in T}h_v},
\label{eq:hook}
\end{equation}
where the hook length ($h_v$) of vertex $v$ is the number of vertices
in the subtree rooted at $v$, including $v$ itself. Notice that the factorial in the numerator of Eq.~\ref{eq:hook} cancels out with the factorial in Eq.~\ref{eq:integration}, simplifying its calculation.

{\bf Probability of a tree given a node ordering.} Given a compatible node ordering,
the probability of generating the tree is
computed by replaying the sublinear
preferential attachment process. Let $v_t$ denote the node introduced at
step $t$, and let $V_{t-1}$ denote the set
of nodes already present.
If node $u\in V_{t-1}$ currently has
$k_u(t)$ direct dependents,
its attachment weight is, for a given $\alpha \in (0,1)$,
\begin{equation}
w(u,\alpha)=
(k_u(t)+1)^\alpha.
\label{eq:weight}
\end{equation}
The probability that the newly introduced
node attaches to its observed parent
$p_t$ is therefore
\begin{equation}
P(v_t\rightarrow p_t\mid V_{t-1},\alpha)
=
\frac{w(p_t,\alpha)}
{\sum_{u\in V_{t-1}}w(u,\alpha)}.
\label{eq:attach}
\end{equation}
The probability of the complete production
history is obtained by multiplying these
attachment probabilities over all
construction steps,
\begin{equation}
P(T\mid o,\alpha)
=
\prod_{t=2}^{n}
P(v_t\rightarrow p_t\mid V_{t-1},\alpha),
\label{eq:history}
\end{equation}
while updating direct dependent counts after
each attachment.

{\bf Monte Carlo estimation.} The number of node orderings compatible with a given tree ($|V(T)|$)
can become extremely large as tree size increases. Exhaustive evaluation
of Eq.~\ref{eq:integration} is therefore
computationally infeasible even for
moderately sized dependency trees. Instead, I estimate it using a Monte Carlo method. The probability in Eq.~\ref{eq:integration} requires averaging jointly over compatible node orderings and over the attachment exponent $\alpha$. For this exponent, I use a uniform prior over the $[0,1]$ interval.

For each Monte Carlo sample $i$, I independently draw a compatible
node ordering $o_i$ uniformly from $V(T)$ and an attachment exponent $
\alpha_i\sim\mathcal{U}_{[0,1]}$, and evaluate $P(T\mid o_i,\alpha_i)$ by replaying the growth process. For $N$ samples, the integrated mean conditional probability is then estimated as
\begin{equation}
\mathbb{E}_{o}[P(T\mid o)]
\approx
\frac{1}{N}
\sum_{i=1}^{N}
P(T\mid o_i,\alpha_i).
\label{eq:integral2}
\end{equation}
Substituting this estimate into Eq.~\ref{eq:integration} gives
\begin{equation}
\widehat P(T)
=
\frac{|V(T)|}{n!}
\frac{1}{N}
\sum_{i=1}^{N}
P(T\mid o_i,\alpha_i).
\label{eq:integral3}
\end{equation}
This Monte Carlo estimator jointly marginalises over uncertainty in
production order and in the sublinear attachment exponent. This estimator is unbiased and converges to the exact expectation
as $N$ increases (see Extended Data Fig.~\ref{fig:convergence}). All analyses use $N=1000$ joint samples
per tree.

\subsection{Generation of random tree controls}

Uniform random-tree baselines for each number of nodes were generated
by uniformly sampling Pr\"ufer sequences\citeMethods{Pruefer:1918} and then uniformly selecting
a root node to determine the orientation of the edges.

Random sPA tree baselines were generated sequentially, starting from a single
root node and adding one node at each step.
Each new node was attached to an existing node $i$ with probability
proportional to $(k_i+1)^\alpha$, where $k_i$ is the current number of direct dependents of node $i$. Trees were grown until they matched the number of nodes of the corresponding attested dependency tree. I used a fixed value of $\alpha=.42$, which was previously found to be optimal for human language sentences.$^{11}$

\subsection{Language-specific probability estimation.}

I converted the dependency trees into phrase-structure trees using a
common procedure.\cite{Moscoso:2025} These trees were built to have parts-of-speech as their leaves (i.e., they were preterminalised).
From the resulting non-empty phrase-structure trees (without any
minimum or maximum number of words), I induced, by maximum likelihood,
a separate PCFG for each language.
I restricted these analyses to languages with at least 10,000 non-empty
dependency trees to ensure that the PCFG estimates were based on
sufficiently large samples, resulting in 34 languages.
The language-specific log probability of each selected sentence was
approximated by the log probability assigned by the corresponding
PCFG to its phrase-structure tree, computed as the sum of the log
probabilities of the grammar rules used in its derivation.

\subsection{Statistical Analyses}

I tested the relationship between the universal sPA and
language-specific PCFG probabilities using a linear mixed-effects
model with log PCFG probability as the dependent variable and log sPA
probability, sentence length, and their interaction as fixed effects.
PCFG and sPA log probabilities were independently standardised to
zero mean and unit variance within each sentence length, and the sentence lengths were centred to have a mean of zero.

I compared two random-effects structures fitting the models by maximum likelihood: a model containing only a random intercept for language,
and a model additionally containing a language-specific random slope
for sPA probability.
Adding the random slope substantially improved model fit
(likelihood-ratio test: $\chi^2(2)=403.23$, $P<.0001$), and produced
lower AIC ($42,157.42$ versus $41,758.19$) and BIC
($42,203.86$ versus $41,820.12$).
I therefore retained the random intercept and random slope and
refitted the final model using restricted maximum likelihood.

The final model showed a strong positive relationship between
length-standardised sPA and PCFG probabilities
($\beta=.171$, 95\% CI $[.142,.199]$, $z=11.701$, $P<.0001$).
Sentence length also had a significant main effect
($\beta=.066$, 95\% CI $[.043,.088]$, $z=5.744$, $P<.0001$), together with a slight interaction between sPA probability and sentence length indicating a slightly weaker effect for longer sentences ($\beta=-.004$, 95\% CI $[-.006,-.002]$, $z=-3.323$,
$P=.001$).

Models were fitted in Python using the \texttt{statsmodels}
package.\citeMethods{Seabold:2010}

\end{methods}
\bibliographyMethods{methods}

\theendnotes

%% Here is the endmatter stuff: Supplementary Info, etc.
%% Use \item's to separate, default label is "Acknowledgements"

\begin{addendum}
 \item I am indebted to Enrique Amig\'o, Michael Anderson, Laurie B. Feldman, Przemek Kubiak, Paul Siewert, and Andreas Vlachos for helpful suggestions.
 \item [Code Availability] Python functions for computing probabilities of dependency trees under the sPA prior, sampling compatible node orderings, and sampling trees from the sPA and uniform models are available at \url{https://github.com/fermosc24/ProbTreesPA}.
 \item[Competing Interests] The author declares that he has no
competing financial interests.
 \item[Correspondence] Correspondence and requests for materials
should be addressed to F. MdP. ~(email: fm611@cst.cam.ac.uk).
\end{addendum}

\clearpage
\renewcommand{\figurename}{Extended Data Figure}
\setcounter{figure}{0}

\begin{figure}[ht]
\centering
\includegraphics[width=.9\linewidth]{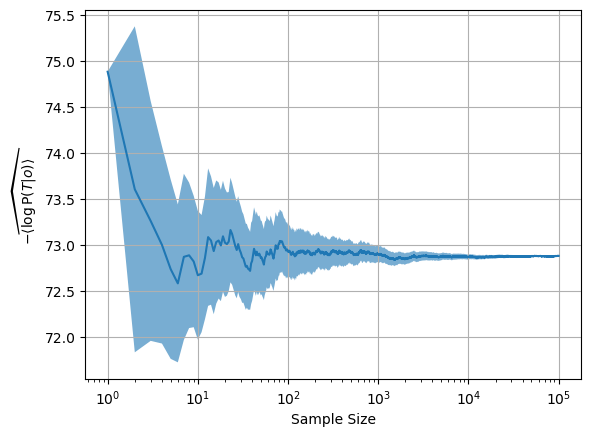}
\caption{{\bf Convergence of the Monte Carlo mean conditional probability estimate.} The figure plots the mean estimated surprisal $-\left<logP(T | o)\right>_{o \in V(T)}$ as a function of the number of sampled histories for a 30-node tree randomly generated by sublinear preferential attachment with $\alpha =0.42$. The shaded region denotes the 95\% confidence interval of the estimate. The horizontal axis is logarithmic.}
\label{fig:convergence}
\end{figure}

%%
%% TABLES
%%
%% If there are any tables, put them here.
%%
\clearpage
\renewcommand{\tablename}{Extended Data Table}

\begin{longtable}{
    >{\raggedright\arraybackslash}p{0.24\textwidth}
    >{\raggedright\arraybackslash}p{0.20\textwidth}
    >{\raggedright\arraybackslash}p{0.25\textwidth}
    r
}
\caption{Languages included in the analysis, ordered by decreasing number of valid available trees.}
\label{tab:languages}\\

\toprule
Language & Family & Group & Number of Valid Trees \\
\midrule
\endfirsthead

\multicolumn{4}{l}{\tablename~\thetable\ -- continued from previous page}\\
\toprule
Language & Family & Group & Number of Valid Trees \\
\midrule
\endhead

\midrule
\multicolumn{4}{r}{Continued on next page}\\
\endfoot

\bottomrule
\endlastfoot

German & Indo-European & West Germanic & 198,830 \\
Japanese & Japonic & Japanese & 120,218 \\
Czech & Indo-European & West Slavic & 115,497 \\
Russian & Indo-European & East Slavic & 101,260 \\
Portuguese & Indo-European & Romance & 67,709 \\
Turkish & Turkic & Oghuz & 63,453 \\
Latin$^{\dagger}$ & Indo-European & Italic (Latino-Faliscan) & 53,303 \\
Icelandic & Indo-European & North Germanic & 52,395 \\
French & Indo-European & Romance & 46,378 \\
Classical Chinese$^{\dagger}$ & Sino-Tibetan & Sinitic & 45,893 \\
English & Indo-European & West Germanic & 41,123 \\
Romanian & Indo-European & Romance & 39,802 \\
Norwegian & Indo-European & North Germanic & 37,679 \\
Italian & Indo-European & Romance & 36,676 \\
Polish & Indo-European & West Slavic & 35,839 \\
Persian & Indo-European & Iranian & 34,685 \\
Spanish & Indo-European & Romance & 34,134 \\
Estonian & Uralic & Finnic & 34,072 \\
Korean & Koreanic & Korean & 33,191 \\
Finnish & Uralic & Finnic & 31,574 \\
Ancient Greek$^{\dagger}$ & Indo-European & Hellenic & 28,975 \\
Arabic & Afro-Asiatic & Semitic & 27,500 \\
Belarusian & Indo-European & East Slavic & 21,060 \\
Old East Slavic$^{\dagger}$ & Indo-European & East Slavic & 19,303 \\
Dutch & Indo-European & West Germanic & 18,354 \\
Hindi & Indo-European & Indo-Aryan & 17,611 \\
Catalan & Indo-European & Romance & 16,500 \\
Old French$^{\dagger}$ & Indo-European & Romance & 15,927 \\
Latvian & Indo-European & Baltic & 15,632 \\
Slovenian & Indo-European & South Slavic & 14,932 \\
Chinese & Sino-Tibetan & Sinitic & 12,300 \\
Swedish & Indo-European & North Germanic & 11,594 \\
Hebrew & Afro-Asiatic & Semitic & 11,015 \\
Bulgarian & Indo-European & South Slavic & 10,451 \\
Croatian & Indo-European & South Slavic & 8,921 \\
Basque & {\em isolate} & -- & 8,691 \\
Slovak & Indo-European & West Slavic & 8,678 \\
Naija & {\em creole} & English-lexifier & 8,619 \\
Indonesian & Austronesian & Malayic & 7,555 \\
Ukrainian & Indo-European & East Slavic & 6,403 \\
Western Armenian & Indo-European & Armenian & 6,205 \\
Pomak & Indo-European & South Slavic & 5,791 \\
Irish & Indo-European & Celtic & 5,773 \\
Old Church Slavonic$^{\dagger}$ & Indo-European & South Slavic & 5,449 \\
Urdu & Indo-European & Indo-Aryan & 5,128 \\
Danish & Indo-European & North Germanic & 5,087 \\
Galician & Indo-European & Romance & 4,899 \\
Gothic$^{\dagger}$ & Indo-European & East Germanic & 4,742 \\
Armenian & Indo-European & Armenian & 4,448 \\
Serbian & Indo-European & South Slavic & 4,363 \\
Scottish Gaelic & Indo-European & Celtic & 4,275 \\
Lithuanian & Indo-European & Baltic & 3,641 \\
Sanskrit$^{\dagger}$ & Indo-European & Indo-Aryan & 3,380 \\
Uyghur & Turkic & Karluk & 3,221 \\
Vietnamese & Austroasiatic & Vietic & 2,886 \\
North Sami & Uralic & Sami & 2,728 \\
Faroese & Indo-European & North Germanic & 2,720 \\
Welsh & Indo-European & Celtic & 2,445 \\
Greek & Indo-European & Hellenic & 2,423 \\
Manx$^{\dagger}$ & Indo-European & Celtic & 2,207 \\
Coptic$^{\dagger}$ & Afro-Asiatic & Egyptian & 2,146 \\
Turkish German & {\em code-switching} & Oghuz / West Germanic & 2,140 \\
Wolof & Atlantic-Congo & Atlantic & 2,076 \\
Afrikaans & Indo-European & West Germanic & 1,929 \\
Maltese & Afro-Asiatic & Semitic & 1,899 \\
Hindi English & {\em code-switching} & Indo-Aryan / West Germanic & 1,894 \\
Akkadian$^{\dagger}$ & Afro-Asiatic & Semitic & 1,844 \\
Hungarian & Uralic & Ugric & 1,780 \\
Erzya & Uralic & Mordvinic & 1,695 \\
Ancient Hebrew$^{\dagger}$ & Afro-Asiatic & Semitic & 1,579 \\
Kiche & Mayan & K'ichean & 1,121 \\
Mbya Guarani & Tupian & Tupi-Guarani & 1,086 \\
Guajajara & Tupian & Tupi-Guarani & 1,017 \\
Amharic & Afro-Asiatic & Semitic & 1,013 \\
Thai & Kra-Dai & Tai & 999 \\
Kazakh & Turkic & Kipchak & 958 \\
Gheg & Indo-European & Albanian & 956 \\
Bambara & Mande & Western Mande & 930 \\
Tamil & Dravidian & South Dravidian & 878 \\
Buryat & Mongolic & Central Mongolic & 844 \\
Cantonese & Sino-Tibetan & Sinitic & 838 \\
Breton & Indo-European & Celtic & 807 \\
Western Sierra Puebla Nahuatl & Uto-Aztecan & Nahuan & 794 \\
Komi Zyrian & Uralic & Permic & 787 \\
Xibe & Tungusic & Jurchenic & 780 \\
Kurmanji & Indo-European & Iranian & 743 \\
Telugu & Dravidian & South-Central Dravidian & 665 \\
Chukchi & Chukotko-Kamchatkan & Chukotkan & 653 \\
Upper Sorbian & Indo-European & West Slavic & 638 \\
Zaar & Afro-Asiatic & Chadic & 581 \\
Tupinamba$^{\dagger}$ & Tupian & Tupi-Guarani & 429 \\
Marathi & Indo-European & Indo-Aryan & 393 \\
Frisian Dutch & Indo-European & West Germanic & 385 \\
Moksha & Uralic & Mordvinic & 350 \\
Bhojpuri & Indo-European & Indo-Aryan & 349 \\
Yoruba & Atlantic-Congo & Volta-Niger & 315 \\
Karo & Tupian & Ramarama & 297 \\
Ligurian & Indo-European & Romance & 297 \\
Yupik & Eskimo-Aleut & Yupik & 284 \\
Kangri & Indo-European & Indo-Aryan & 281 \\
Yakut & Turkic & Siberian Turkic & 237 \\
Karelian & Uralic & Finnic & 220 \\
Skolt Sami & Uralic & Sami & 216 \\
Nheengatu & Tupian & Tupi-Guarani & 178 \\
Tagalog & Austronesian & Philippine & 178 \\
Swedish Sign Language & {\em sign language} & -- & 169 \\
Cebuano & Austronesian & Philippine & 150 \\
Tatar & Turkic & Kipchak & 146 \\
Teko & Tupian & Tupi-Guarani & 132 \\
Hittite$^{\dagger}$ & Indo-European & Anatolian & 130 \\
Javanese & Austronesian & Malayo-Polynesian & 123 \\
Livvi & Uralic & Finnic & 119 \\
Munduruku & Tupian & Munduruku & 113 \\
Swiss German & Indo-European & West Germanic & 100 \\
Sinhala & Indo-European & Indo-Aryan & 100 \\
Apurina & Arawakan & Purus & 98 \\
Low Saxon & Indo-European & West Germanic & 93 \\
Akuntsu & Tupian & Tupari & 93 \\
Abaza & Northwest Caucasian & Abkhaz-Abaza & 86 \\
Komi Permyak & Uralic & Permic & 86 \\
South Levantine Arabic & Afro-Asiatic & Semitic & 84 \\
Umbrian$^{\dagger}$ & Indo-European & Italic (Sabellic) & 80 \\
Albanian & Indo-European & Albanian & 60 \\
Beja & Afro-Asiatic & Cushitic & 52 \\
Kaapor & Tupian & Tupi-Guarani & 48 \\
Warlpiri & Pama-Nyungan & Ngumpin-Yapa & 47 \\
Assyrian & Afro-Asiatic & Semitic (Aramaic) & 45 \\
Bengali & Indo-European & Indo-Aryan & 36 \\
Makurap & Tupian & Tupari & 24 \\
Malayalam & Dravidian & South Dravidian & 23 \\
Guarani & Tupian & Tupi-Guarani & 22 \\
Old Turkish$^{\dagger}$ & Turkic & Common Turkic & 16 \\
Xavante & Macro-Je & Je & 16 \\
Madi & Arawan & Madi & 11 \\
Nayini & Indo-European & Iranian & 10 \\
Khunsari & Indo-European & Iranian & 9 \\
Soi & Indo-European & Iranian & 7 \\
Neapolitan & Indo-European & Romance & 1 \\

\multicolumn{4}{l}{$^{\dagger}$Extinct language.}\\

\end{longtable}

\end{document}